\documentclass[runningheads]{llncs}
\usepackage[T1]{fontenc}
\usepackage{changepage}
\usepackage{graphicx}
\usepackage{booktabs}

\usepackage{comment}
\usepackage{amsmath}
\usepackage{subfig}
\usepackage{caption}
\usepackage[noadjust]{cite} 
\usepackage{xcolor} 
\usepackage{listings} 
\usepackage{algorithm}
\usepackage{algpseudocode}
\usepackage{tcolorbox} 
\usepackage{subfig} 

\begin{document}
\title{LLM-Foraging: Large Language Models for Decentralized Swarm Robot Foraging}
\titlerunning{LLMs for Swarm Robot Foraging}
\author{Peihan Li\inst{1} \and Joanna Gutierrez\inst{2} \and Fabian Hernandez\inst{2} \and Qi Lu\inst{2} \and Lifeng Zhou\inst{1}}
\authorrunning{P. Li et al.}
%
\institute{Drexel University, Philadelphia, PA 19104, USA \\
\email{\{pl525, lz457\}@drexel.edu}\and
The University of Texas Rio Grande Valley, Edinburg, Texas 78541, USA \\
\email{\{joanna.gutierrez03,fabian.hernandez04,qi.lu\}@utrgv.edu}
}
\maketitle              
\begin{abstract}
Swarm foraging algorithms, such as the central-place foraging algorithm (CPFA), typically rely on offline parameter optimization using genetic algorithms (GA) or reinforcement learning, yielding policies tightly coupled to a specific combination of team size, arena size, and resource distribution. When deployment conditions change, performance degrades, and retraining is computationally expensive. We propose LLM-Foraging, a decentralized swarm controller that augments the CPFA state machine with a large language model (LLM) tactical decision-maker at three structured decision points, namely post-deposit, central-zone arrival, and search starvation. Each robot runs its own LLM client and queries it using only locally observable state, while the existing CPFA motion and sensing stack executes the selected action. Because the LLM serves as a general decision policy rather than parameters fitted to a single configuration, the controller is training-free at deployment and transfers across configurations without re-optimization. We evaluate LLM-Foraging in Gazebo with TurtleBot3 robots across 36 configurations spanning team sizes of 4 to 10 robots, arena sizes from $6\times6$ to $10\times10$ meters, and three resource distributions (clustered, powerlaw, random). LLM-Foraging collects more resources than the GA-tuned CPFA baseline across the evaluated configurations and is more consistent, a property that the GA's single-configuration tuning does not transfer.

\end{abstract}
%
%

\section{Introduction}

Swarm robotics is inspired by the fascinating collective behaviors exhibited in social organisms such as ant colonies~\cite{Crist1991,pherm-bio2}, honey bees, bird flocks~\cite{2010StarlingFlocks}, and schools of fish~\cite{2019FishSchool}. Individual agents follow simple rules and cooperate with other agents without a central leader, which demonstrates collective emerging behavior in these natural systems. Contemporary research encompasses a range of collective emerging behaviors, including task allocation~\cite{acbba2020,2025MRSAuctionTaskAlloc}, aggregation~\cite{cue-based1,cue-based2}, object sorting~\cite{sorting0,sorting1}, and foraging~\cite{2018foragingQuality,Lu2018DynamicDepots, Jin2020ForagingDRL,kaminka2025heterogeneous}.

In this work, we focus on a promising area of research in the foraging of swarm robots. In the foraging task, robots search for multiple unknown objects or resources (e.g., minerals or survivors) in a large unknown area and deliver them to a specified collection zone (e.g., warehouses or hospitals). In recent work, efficient foraging algorithms were developed for robot swarms~\cite{BeyondPherom2015,2016IROSMPFA, Lu2018DynamicDepots, MatthewDDSA2016, LuICRA2020, RobotChain2021, RobotChainNetwork2022}. The goal is to coordinate a group of robots to search for as many resources as possible in a limited time window. Most of the work focuses on developing stochastic foraging algorithms~\cite{BeyondPherom2015,2016IROSMPFA, Lu2018DynamicDepots} since they are more robust and adaptive than the deterministic foraging algorithm~\cite{MatthewDDSA2016,Luna2023AMDDSA} in the dynamic real-world environment~\cite{LuICRA2020}. The foraging task has various real-world applications, such as search and rescue, toxic waste cleanup, humanitarian demining, and agricultural harvesting~\cite{2013SwarmEngPersp, 2023MultRobotAgri,2025RobotMappingHarvest}. 

The performance of stochastic foraging robot algorithms can be optimized using AI techniques~\cite{Ferrante2013GESwarm, BeyondPherom2015, 2016IROSMPFA, Jin2020ForagingDRL}. Despite the efficiency of traditional training methods, the resulting policies are tied to the number of robots and environments used for training. The CPFA~\cite{BeyondPherom2015} and its variants, such as MPFA~\cite{2016IROSMPFA, Lu2018DynamicDepots}, rely on offline parameter optimization, typically through genetic algorithms, to discover effective foraging strategies for a given combination of swarm size, arena size, and resource distribution. Learning-based approaches such as deep reinforcement learning~\cite{Jin2020ForagingDRL} face the same limitation: the learned policy is fit to the distribution of environments seen during training. When any of these conditions change at deployment time, the trained parameters no longer match the new regime, and performance largely degrades. Retraining for each new configuration requires access to the target environment and substantial computational resources and time, which is impractical when deployment conditions are unknown in advance or change during operation.

Large Language Models (LLMs) have been shown to exhibit problem-solving behavior on tasks they were not explicitly trained for, using only natural-language prompts that describe the situation and the expected response format~\cite{kojima2022large,wei2022chain,yao2022react}.
Recent work in robotics has applied LLMs as high-level planners or policy generators, primarily in single-robot manipulation and navigation~\cite{saycan2022arxiv,liang2023code,huang2023inner}.
Within multi-robot systems, LLMs have been applied across task allocation and planning~\cite{li2025large,kannan2024smart}, as in-the-loop optimizers for real-time target tracking under uncertainty~\cite{wu2025hierarchical}, and for decentralized formation control and flocking~\cite{li2025llm}.

We propose LLM-Foraging, a decentralized swarm-foraging controller.
Each robot runs its own LLM client and queries it only at three decision points of the CPFA state machine, using only locally observable state.
The three points correspond to a successful deposit at the central collection zone, an empty-handed return, and after a fixed period of unsuccessful search.
At each point, the LLM selects from the existing CPFA actions, and the CPFA motion stack executes the choice.
Unlike per-step LLM controllers, such as the NetLogo ant agents of Jimenez-Romero et al.~\cite{jimenez2025multi}, and mission-level LLM planners, such as SMART-LLM~\cite{kannan2024smart}, LLM-Foraging confines the LLM to these three points while leaving the rest of the CPFA controller intact.
Because the LLM is a general decision policy rather than parameters fitted to one configuration, the controller is training-free at deployment and transfers across team size, arena size, and resource distribution without re-optimization.
We evaluate LLM-Foraging in ROS-Gazebo using TurtleBot3 robots across 36 configurations.
It deposits more resources than the GA-tuned CPFA baseline in 33 of the cells, with the largest gains in clustered and powerlaw layouts, where site fidelity and pheromone trails pay off.

The remainder of this article is organized as follows. Section~\ref{sec-related} reviews related works in swarm robots foraging and LLMs in multi-robot systems. Section~\ref{sec-cpfa} covers the background of the CPFA algorithm. Section~\ref{sec-methods} presents the framework of our proposed LLM-Foraging. Section~\ref{sec-exp} explains the experimental setup and demonstrates the results. Section~\ref{sec-conclusion} shows our conclusion and limitations.


\section{Related Works}
\label{sec-related}

Several traditional training methods exist for optimizing stochastic foraging algorithms for robot swarms. 
Zaman et al.~\cite{Zaman2025NEAT} apply NEAT (NeuroEvolution of Augmented Topologies) to evolve adaptive swarm foraging behaviors in obstacle-filled, unknown environments. The proposed P-NeatFA strategy uses reward-based training to improve foraging efficiency, obstacle avoidance, and scalability, significantly outperforming CPFA and NeatFA in simulations across different resource distributions and swarm sizes. 
Jin et al.~\cite{Jin2020ForagingDRL} employ deep reinforcement learning with raw camera images to train a swarm of robots for cooperative foraging. Results show that image-based controllers can successfully achieve collective transport behavior, with improved variants enhancing performance and adaptability.
Song et al.~\cite{Song2020ForagingNN} develop a neural-network-based pheromone model for swarm robotic foraging, where neuron outputs represent pheromone density that diffuses and evaporates over time. Using mathematical optimization and differential equations for task allocation, the authors determine key parameters for cooperative foraging and show, through simulations, that the approach effectively improves swarm foraging performance across different scenarios.
Just et al.~\cite{Just2018ForagingNN} propose a flexible swarm foraging algorithm in which each robot dynamically selects among evolved foraging strategies using only local sensing. Decentralized decision-making improves adaptability and enables the swarm to outperform homogeneous swarms across diverse resource distributions without prior tuning.
Although these methods improve foraging performance, they commonly rely on offline training or parameter optimization under specific environmental assumptions. As a result, the optimized policy may not transfer well when the team size, arena size, or resource distribution changes, motivating decision-making mechanisms that can adapt during deployment without a new round of offline optimization.

Recent work has begun to integrate LLMs into multi-robot systems across a range of roles, surveyed in~\cite{li2025large}.
At the task-planning level, SMART-LLM~\cite{kannan2024smart} decomposes natural-language mission descriptions into coordinated assignments for a multi-robot team, and RoCo~\cite{mandi2024roco} applies a dialectic collaboration scheme to multi-arm manipulation.
For multi-robot navigation, Co-NavGPT~\cite{yu2026conavgpt} employs a vision-language model as a global planner to allocate unexplored frontiers to individual robots based on merged semantic maps.
For decentralized formation control, LLM-Flock~\cite{li2025llm} and FlockGPT~\cite{lykov2024flockgpt} both apply LLMs to flocking, with LLM-Flock introducing an influence-based consensus protocol and a two-layer collision-avoidance mechanism to stabilize behavior across agents.
For real-time decision loops, Wu et al.~\cite{wu2025hierarchical} use a hierarchical LLM architecture as an in-the-loop optimizer that adjusts parameters of a bi-level task allocation problem rather than generating robot control directly.
Within foraging specifically, Jimenez-Romero et al.~\cite{jimenez2025multi} replace the hard-coded controllers of NetLogo ant agents with LLM-driven prompts operating over continuous pheromone gradients and nest-scent fields.

Integrating LLMs into multi-robot systems raises architectural choices along a centralized-to-decentralized spectrum~\cite{li2025large}.
In a centralized architecture, a single LLM instance observes the state of all robots and produces coordinated plans or actions for the team~\cite{kannan2024smart}.
This simplifies coordination because the LLM can reason over the entire team, enforce consistency, and produce a unified plan, but it introduces scalability limits, communication overhead, latency that may grow with team size, and sensitivity to a single point of failure.
In a decentralized architecture, each robot runs its own LLM instance and makes decisions from locally available information~\cite{li2025llm}.
This avoids reliance on a central decision-maker, but shifts the coordination burden to local decisions and indirect signals.
Hybrid architectures combine the two by placing a coordinating LLM above per-robot LLM agents to balance global coherence against communication cost~\cite{chen2024scalable}.
On warehouse task benchmarks, hybrid architectures have been reported to outperform both pure centralized and pure decentralized configurations once team size grows beyond a handful of robots~\cite{chen2024scalable}.

Beyond architecture, LLMs in multi-robot systems can be deployed at different levels of the control hierarchy, from high-level task allocation and planning to mid-level motion planning and navigation, and low-level action or motor-command generation~\cite{li2025large}.
Prior LLM-foraging work places the LLM close to the low-level action loop, where it directly generates each action step~\cite {jimenez2025multi}.
In contrast, we embed the LLM at structured decision points inside the CPFA state machine rather than using it to generate complete plans, low-level control commands, or replace the entire agent controller.
At each such point, the LLM selects among existing CPFA actions based only on the robot's local state.
This positions the LLM as an intermediate tactical decision-maker that augments an existing swarm-foraging controller while leaving the CPFA motion, sensing, pheromone, and fallback mechanisms intact.

For swarm foraging, the architectural choice is also constrained by the underlying algorithm rather than by performance optimization alone.
Classical swarm foraging algorithms such as CPFA~\cite{BeyondPherom2015} assume robots operate based on local sensing, with indirect coordination through pheromone deposition and collocation.
We therefore adopt a fully decentralized architecture in which each robot runs its own LLM instance and queries it only with local information, matching the decentralized assumptions of the underlying CPFA controller.
Our foraging setting involves individual food items that deplete on pickup, virtual pheromone waypoints, and site fidelity within a central-place framework, which distinguishes it from the continuous-gradient, stateless-agent NetLogo setting of~\cite{jimenez2025multi}.
To our knowledge, no prior work applies an LLM at the decision points of a classical central-place foraging controller in a decentralized swarm setting.

\section{The Background of CPFA}
\label{sec-cpfa}
\begin{figure}[htbp!]
    \centering
    \includegraphics[width=0.8\textwidth]{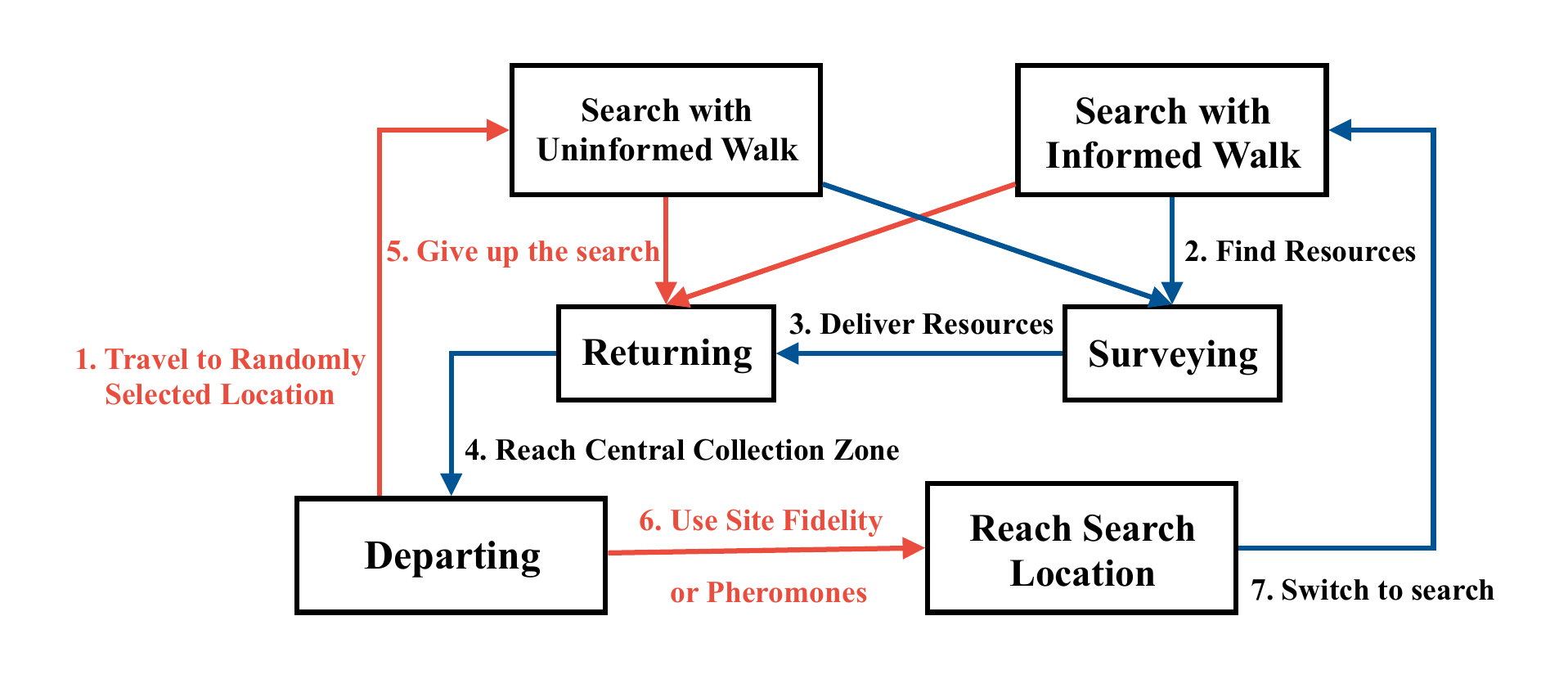}
    \vspace{-2mm}
    \caption{The flow chart of an individual robot's behavior and states in the CPFA. The red links show the decision point replaced by LLM queries in our proposed LLM-Foraging approach (see Section~\ref{sec-methods}).}
    \label{fig_states}
\end{figure}

The CPFA algorithm is inspired by the foraging behavior of ant colonies~\cite{BeyondPherom2015}. Its main purpose is to coordinate a swarm of robots to autonomously forage for resources (e.g., minerals or survivors) in a large unknown environment and then deliver them to the central collection zone. The behavior of an individual robot in the CPFA foraging algorithm is shown in Fig.~\ref{fig_states}. 
\vspace{-5mm}

Initially, a robot leaves the central collection zone and randomly searches for resources (Step 1). If it finds resources (Step 2), it will create a virtual pheromone waypoint based on the detected resource density around that location. Upon returning to the center (Step 3), it shares this waypoint with the central server in the collection zone. Once it arrives at the center (Step 4), it drops off the collected resource and can access other pheromone information on the central server. It can decide where to go based on the shared pheromone or site-fidelity (the last location where it found resources) information in the central server (Step 6), or search for resources again at random. Upon reaching the location, it searches for resources using the informed random walk (Step 7). If it can not find resources, it gives up the search and returns to the center (Step 5).



A set of real-valued parameters specifying the individual robot controllers. Our robots mimic the behaviors of seed-harvester ants, which have evolved over millions of years. We encode the set of seven real-valued parameters of CPFA for the GA training (see
Table 1). The parameters specify the probabilistic decision-making from states to states in movement, sensing, and communication (see Fig.~\ref{fig_states}). The detailed explanation of the parameters is provided in~\cite{Lu2018DynamicDepots,BeyondPherom2015}, and the configuration of the GA training is described in Section~\ref{sec-exp}.

\begin{table}[!htbp]
\centering
\caption{CPFA Parameters for decision making and GA training}
\label{tab:cpfa_params}
\scriptsize
\renewcommand{\arraystretch}{0.8}
\setlength{\tabcolsep}{13pt}
\begin{tabular}{llll}
\toprule
\textbf{Parameter} & \textbf{Description} & \textbf{Range} & \\
\midrule
 $p_s$   &   Prob.\ of switching to searching    & $\mathcal{U}(0,\ 1)$  \\ [4pt]
 
 $p_r$  &    Prob.\ of returning to the central collection zone      & $\mathcal{U}(0,\ 1)$      \\ [4pt]
 
$\rho_u$   & Uninformed search variation       & $\mathcal{U}(0,\ 4\pi)$    \\ [4pt]

 $\lambda_i$  &  Rate of informed search decay   & $\mathrm{exp}(5)$  \\ [4pt]
 
$\lambda_f$ &  Rate of site fidelity   &   $\mathcal{U}(0,\ 20)$   \\ [4pt]

$\lambda_{lp}$ & Rate of laying pheromone & $\mathcal{U}(0,\ 20)$  \\ [4pt]

 $\lambda_d$   &  Rate of pheromone decay   & $\mathrm{exp}(10)$  \\ [4pt]
\bottomrule
\end{tabular}
\end{table}

\section{LLM-Foraing}
\label{sec-methods}

We present LLM-Foraging, an LLM-augmented CPFA state decision-maker at three decision points.
Each robot runs a local copy of the state machine and queries its own LLM using only locally observable states.
The LLM's output is a discrete action drawn from an event-specific whitelist, and the state machine executes the selected action using the existing CPFA motion and sensing stack.
Algorithm~\ref{alg:llm_cpfa} summarizes the hybrid controller for a single robot.

\begin{algorithm}
\caption{LLM-Foraging CPFA Controller (per robot) \\ 
$c$ is the local resource-density count at the last pickup, and $f$ is the fidelity flag (set on pickup, cleared on give-up). \textsc{sf}, \textsc{p}, \textsc{u} abbreviate \textsc{use\_site\_fidelity}, \textsc{follow\_pheromone}, \textsc{uninformed\_search}.}
\label{alg:llm_cpfa}
\small
\begin{algorithmic}[1]
\State Disperse from the central collection zone to a random location; $f \gets \mathrm{false}$
\While{experiment running}
  \State Conduct correlated random walk (informed or uninformed)
  \If{resource found}
    \State Collect resource; count $c$ near $l_f$; $f \gets \mathrm{true}$
    \State Return to the central collection zone with resource
    \If{$f$ and $\mathrm{CDF}_{\mathrm{Pois}}(c, \lambda_{lp}) > U(0,1)$}
      \State Lay pheromone trail to $l_f$
    \EndIf
    \State $a \gets \textsc{LLM}(\mathrm{POST\_DEPOSIT})$ \Comment{replaces CPFA cascade}
    \State Execute $a \in \{\textsc{sf}, \textsc{p}, \textsc{u}\}$
  \ElsIf{elapsed search $\geq T_s$, re-eval every $T_q$}
    \State $a \gets \textsc{LLM}(\mathrm{SEARCH\_STARVATION})$ \Comment{replaces $p_r$}
    \If{$a = \textsc{return\_for\_info}$}
      \State Return to the central collection zone empty-handed; $f \gets \mathrm{false}$
      \State $a \gets \textsc{LLM}(\mathrm{CENTRAL\_ZONE\_ARRIVAL})$ \Comment{expands 2-way cascade}
      \State Execute $a \in \{\textsc{sf}, \textsc{p}, \textsc{u}\}$
    \EndIf
  \EndIf
\EndWhile
\end{algorithmic}
\end{algorithm}

Figure~\ref{fig_states} marks the three transitions of the CPFA state machine at which an LLM is queried, marked in red.
The \emph{post-deposit} decision fires when a robot has just deposited a resource at the central collection zone.
In vanilla CPFA~\cite{BeyondPherom2015}, the next strategy is determined by a Poisson-CDF cascade in which site fidelity is selected with probability $\mathrm{CDF}_{\mathrm{Pois}}(c, \lambda_{sf})$ based on the last-observed local resource density $c$, conditional on a fresh pickup being available.
If site fidelity is not selected, a pheromone waypoint is followed when one is available, and a new uninformed search location is chosen.
We replace this cascade with a single LLM query whose response selects among \{\texttt{use\_site\_fidelity}, \texttt{follow\_pheromone}, \texttt{uninformed\_search}\}.

The \emph{central-zone-arrival} decision fires when a robot returns to the central collection zone without a resource, after giving up the search.
In vanilla CPFA, the same Poisson-CDF cascade runs at the central collection zone arrival, but because the site-fidelity flag is cleared on give-up, the site-fidelity branch is effectively disabled, and the decision reduces to a two-way cascade (pheromone if available, otherwise a new random location).
We replace this reduced cascade with an LLM query over the full three-action set.
Allowing the LLM to select \textsc{use\_site\_fidelity} here allows it to revisit a remembered pickup location even after an unsuccessful search, an option that vanilla CPFA does not provide at this time.

The \emph{search-starvation} decision fires when a robot has been searching unsuccessfully for $T_s = 60$\,s and is re-evaluated every $T_q = 30$\,s thereafter until the robot either finds a resource or returns to the central collection zone.
Its action space is \{\texttt{continue\_search}, \texttt{return\_for\_info}\}.
In vanilla CPFA, the analogous mechanism is the per-waypoint give-up probability $p_r$ evolved by the GA, where at each search-target waypoint the robot returns to the central collection zone with probability $p_r$ and continues searching otherwise.
Our LLM-based decision replaces this per-waypoint stochastic check with a time-triggered LLM query that can condition on the elapsed search time and the surrounding pheromone context.

\noindent \textbf{Query structure}.
At each decision point, the robot assembles a JSON prompt containing its local state and sends it to the LLM.
The prompt includes the robot identifier, the event type, the elapsed time since the last pickup, the last pickup location, a summary of currently active pheromone waypoints, and the action whitelist for this event.
The LLM returns a JSON object with two fields, \texttt{action} and \texttt{rationale}.
The \texttt{action} field is validated against the event-specific whitelist, while the \texttt{rationale} field is a natural-language justification logged for later analysis that does not affect execution.

On timeout, parse error, or an out-of-whitelist action, the controller falls back to the CPFA parameter-driven choice for that event type. Fig.~\ref{fig:cluster-prompt} shows a concrete prompt and response for a post-deposit decision.

\noindent \textbf{Decentralized implementation}.
Decentralization is preserved at the implementation level.
Each robot's controller maintains a dedicated LLM client and invokes it only at the three decision points described above, using only the information available to that robot.
No messages are exchanged between per-robot LLMs, and no shared LLM context is maintained across robots, so coordination occurs only through the existing pheromone field, as in vanilla CPFA.
The framework makes no assumption about the specific LLM used.
Any model capable of producing the JSON response described above can be substituted without changes to the state machine or the fallback path.

\section{Experiments}
\label{sec-exp}

We evaluate the LLM-Foraging CPFA policy against the GA-tuned CPFA baseline by running every combination of robot team size $\in \{4, 6, 8, 10\}$, squared arena width and length $\in \{6, 8, 10\}$\,m, and resource distribution $\in$ \{\emph{powerlaw}, \emph{clustered}, \emph{random}\}, where we adopt the resource distribution setting from~\cite{BeyondPherom2015}.
This yields 36 experimental combinations per policy as shown in Table~\ref{tab_exp}. We run 10 trials per combination, and each runs 1200 simulated seconds (20 minutes), for a total of 360 trials per policy. Please find the video demonstration in this YouTube link\footnote{\url{https://tinyurl.com/DARSLLMForaging}} and our codebase on GitHub\footnote{\url{https://tinyurl.com/LLM-Foraging-code}}.

\vspace{-3mm}
\begin{table}[!htbp]
\centering
\caption{Experimental Configurations}
\label{tab_exp}
\begin{tabular}{|l|ccc|}
\hline
Arena size (m)  & \multicolumn{1}{c|}{6 $\times$ 6} & \multicolumn{1}{c|}{\textbf{8 $\times$ 8}} & 10 $\times$ 10 \\ \hline
\# of resources & \multicolumn{1}{c|}{64}  & \multicolumn{1}{c|}{\textbf{128}}     & 256     \\ \hline
\# of robots    & \multicolumn{3}{c|}{4, \textbf{6}, 8, 10} \\ \hline
Resource dist.  & \multicolumn{3}{c|}{Clustered, \textbf{Powerlaw}, Random}                                                                            \\ \hline
\end{tabular}
\end{table}

\begin{figure}[!htbp]
    \subfloat[Clustered]{\label{fig:gazebo-cluster}\includegraphics[width=0.32\textwidth]{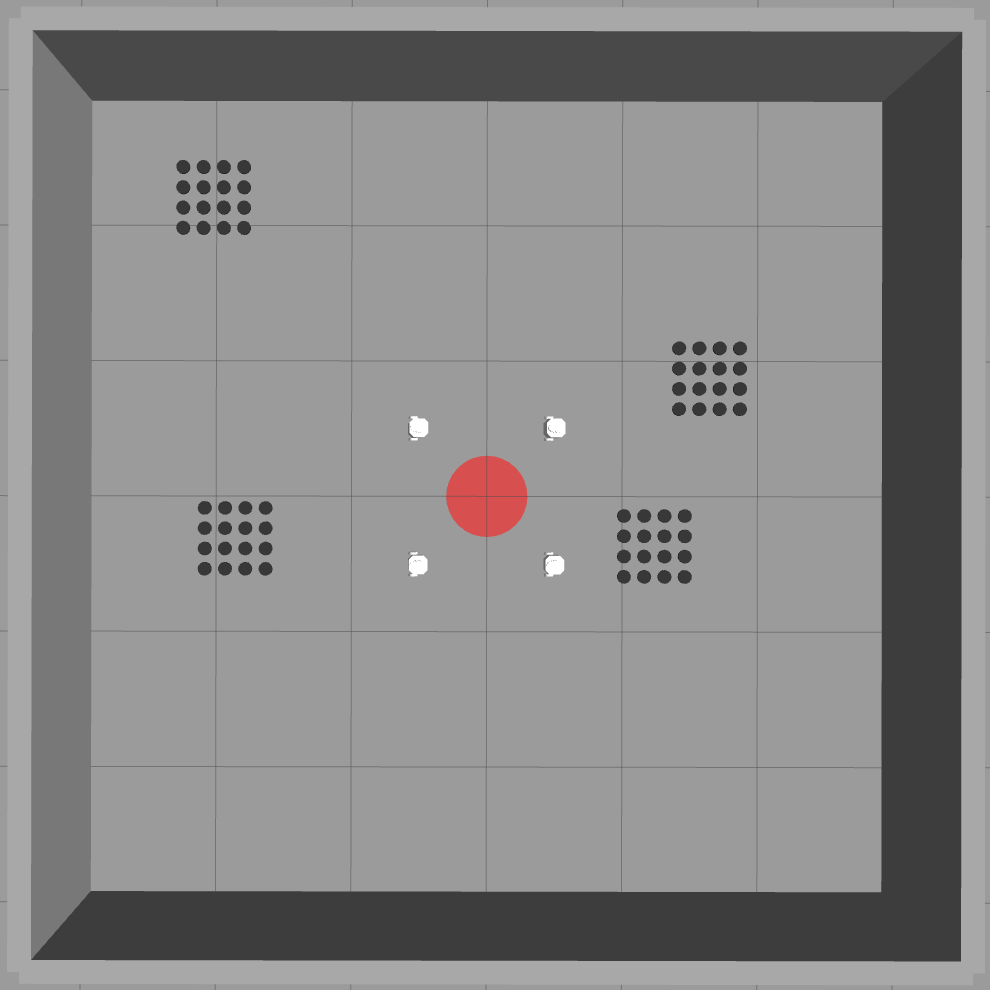}}
    \subfloat[Powerlaw]{\label{fig:gazebo-power}\includegraphics[width=0.32\textwidth]{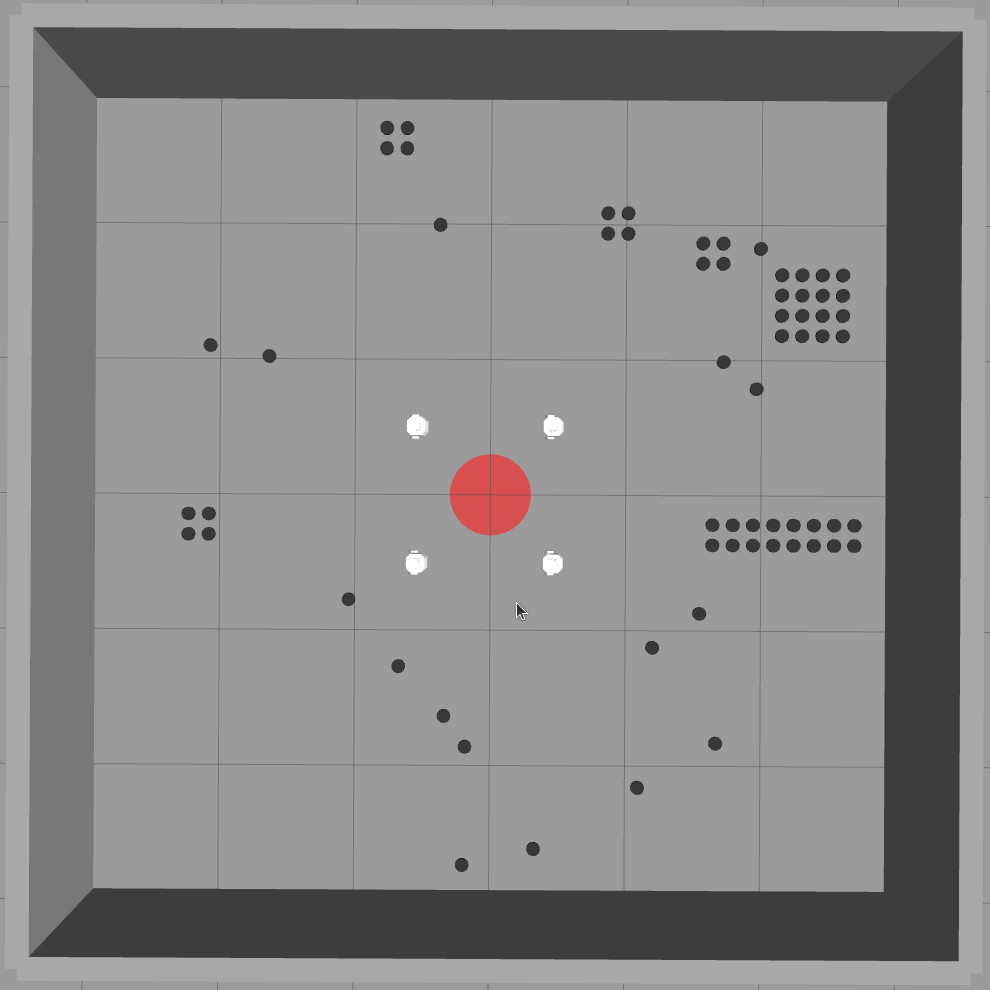}}
    \subfloat[Random]{\label{fig:gazebo-random}\includegraphics[width=0.32\textwidth]{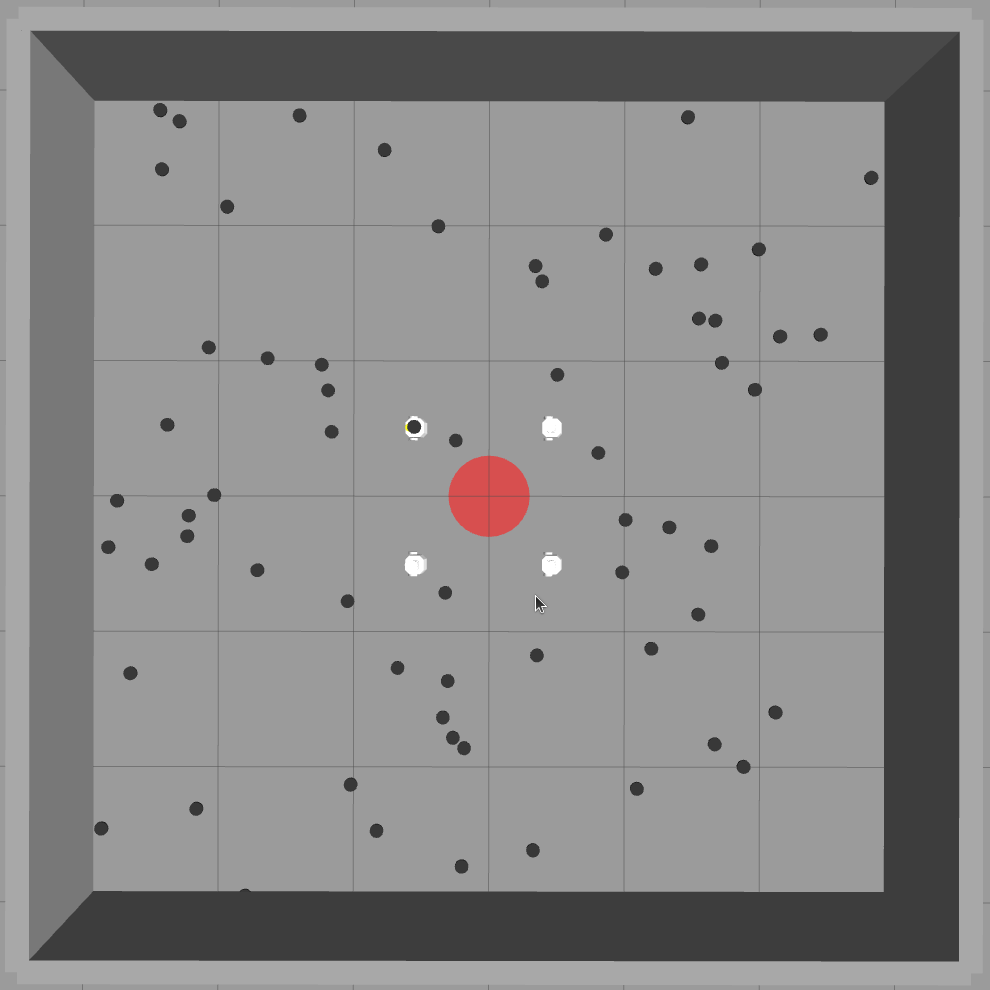}}

    \caption{Initial configuration of the Gazebo arena under each of the three resource distributions. In each panel, the red disk at the center is the central collection zone, the four white markers around it are TurtleBot3 robots at their starting positions, the black disks are resources, and the arena is bounded by four walls.}
    \label{fig:gazebo-distribution}
\end{figure}

The three resource distributions differ in their resource arrangements.
The \emph{powerlaw} distribution, as shown in Fig.~\ref{fig:gazebo-power}, seeds resources into clusters whose sizes decay according to a powerlaw rank schedule (rank parameter~4), producing a heavy-tailed mixture of dense pockets and sparse fringes.
The \emph{clustered} distribution, as shown in Fig.~\ref{fig:gazebo-cluster}, partitions resources into four equally-sized clusters ($8{\times}8$ footprint in cluster-local units). Then, the four clusters are uniformly distributed in the arena.
The \emph{random} distribution, as shown in Fig.~\ref{fig:gazebo-random}, draws resource positions uniformly at random across the arena.

The number of resources scales with arena size to hold density roughly constant, with 64, 128, and 256 items placed in the $6{\times}6$, $8{\times}8$, and $10{\times}10$\,m arenas respectively.
Trials within a set draw from random seed lists, so the two policies face random resource layouts in comparisons.
The primary outcome metric is the number of resources successfully deposited at the central collection zone during a trial.

\subsection{GA-tuned CPFA Baseline}
The numbers and the word \emph{powerlaw} in bold are the configuration for the CPFA GA training. We trained the seven parameters with 6 robots, 128 resources in the powerlaw distribution, and the squared arena size is 8$\times$8 meters. We run 10 trials per model (the GA generates 7 parameters), and each trial runs for 12 minutes. We evaluate the trained model's flexibility across different configurations (the number of robots and resources, resource distributions, and arena sizes), as shown in Table~\ref{tab_exp}.

The computational time $T$ of the GA scales with the trial duration, the number of trials, the population size, and the number of generations. This can be represented through:
\begin{equation}
    T = t_{\text{eval}} \times N_{\text{trials}} \times P \times G
    \label{eq:ga_eq}
\end{equation}
where $t_{\text{eval}}$ is the evaluation time per trial in minutes, $N_{\text{trials}}$ is the number of trials per genome, $P$ the population size per generation, and $G$ the number of generations.

For $t_{\text{eval}}$=12, $N_{\text{trials}}$=10, $P$=10 , $G$=30: 
\begin{equation}
T = 12 \times 10 \times 10 \times 30 = 36{,}000 \text{ min} = 600 \text{ hrs}
\label{eq:ga_cost}
\end{equation}

All experiments run in Gazebo Classic~\cite{koenig2004design} under ROS 2 Humble~\cite{macenski2022ros2}, with each robot modeled as a TurtleBot3 Burger~\cite{amsters2019turtlebot}.
Robots detect food within a 0.3\,m pickup radius and exchange virtual pheromone waypoints through a shared pheromone manager, as in the CPFA (see Section~\ref{sec-cpfa}).
The motion layer uses a continuous-drive controller with linear and angular velocities of 0.3\,m/s and 1.0\,rad/s, respectively, and an ARGoS-style pairwise yield for inter-robot collision avoidance.

\subsection{LLM-Foraging Setup}

For the LLM-Foraging policy, we use OpenAI's \texttt{gpt-5-mini}~\cite{singh2025openai} via the Responses API with \texttt{reasoning\_effort=low}, \texttt{max\_output\_tokens=1024}, and a 30\,s per-call timeout.
The \emph{search-starvation} event first fires after $T_s = 60$\,s of unsuccessful search and is re-evaluated every $T_q = 30$\,s until the robot either finds food or returns to the central collection zone.
On any timeout, parse error, or out-of-whitelist action, the controller falls back to the CPFA parameter-driven choice for that event, so the controller never stalls when a call fails.


\subsection{Qualitative Results}
\label{sec-results}

\begin{figure}
    \centering
    \includegraphics[width=\textwidth]{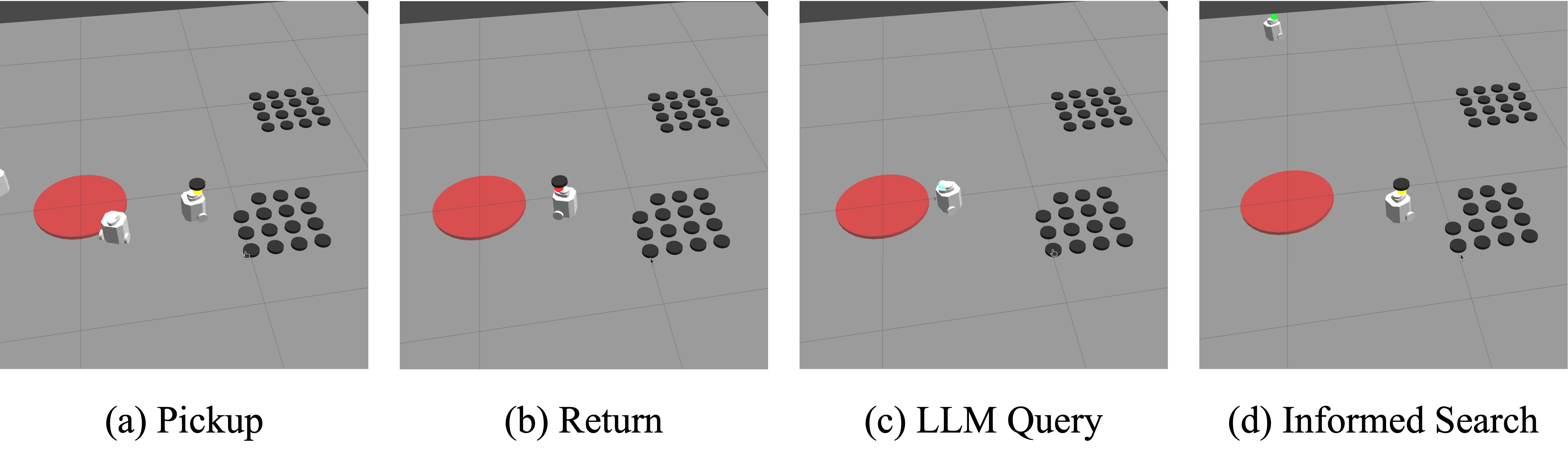}  
    \vspace{-0.7cm}
    \caption{Snapshots from Gazebo simulation demonstrating the process of robot picking up and depositing resources, and querying LLM for the next stage to continue the CPFA mission.}
    \label{fig:cluster-example}
\end{figure}

\begin{figure}[htbp!]
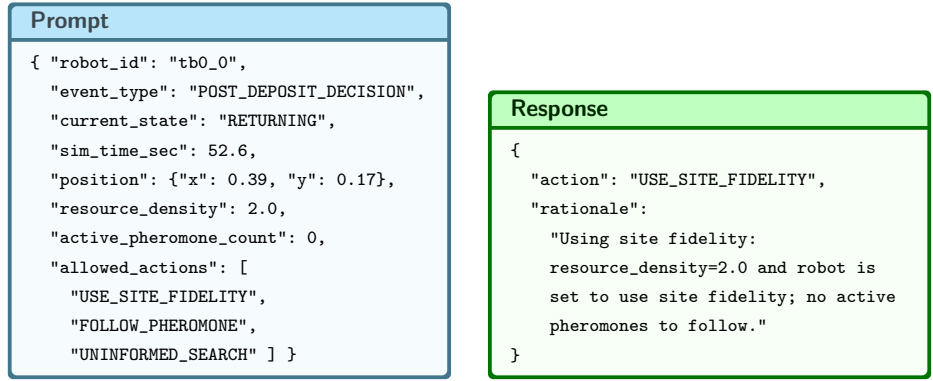
                                        
  \centering           
  \begin{minipage}[t]{0.48\textwidth}
  \begin{tcolorbox}[
    title={\footnotesize\textsf{\textbf{Prompt}}},
    colback=cyan!3, colframe=cyan!45!black,
    colbacktitle=cyan!25, coltitle=cyan!15!black,
    boxsep=2pt, left=5pt, right=5pt, top=3pt, bottom=3pt,
    arc=2pt, outer arc=2pt
  ]                                                                                                                                                    
  {\ttfamily\scriptsize
  \{
  \hspace*{0em}"robot\_id": "tb0\_0",\\
    \hspace*{1em}"event\_type": "POST\_DEPOSIT\_DECISION",\\
  \hspace*{1em}"current\_state": "RETURNING",\\
  \hspace*{1em}"sim\_time\_sec": 52.6,\\                                                                                                               
  \hspace*{1em}"position": \{"x": 0.39, "y": 0.17\},\\                                                                                                 
  \hspace*{1em}"resource\_density": 2.0,\\                                                                                                             
  \hspace*{1em}"active\_pheromone\_count": 0,\\
  \hspace*{1em}"allowed\_actions": [\\
  \hspace*{2em}"USE\_SITE\_FIDELITY",\\                                                                                                                
  \hspace*{2em}"FOLLOW\_PHEROMONE",\\                                                                                                                  
  \hspace*{2em}"UNINFORMED\_SEARCH" ]
  \}              
  }                                                                                                                                                    
  \end{tcolorbox} 
  \end{minipage}\hfill%
  \begin{minipage}[t]{0.48\textwidth}
  \begin{tcolorbox}[                                                                                                                                   
    title={\footnotesize\textsf{\textbf{Response}}},
    colback=green!3, colframe=green!45!black,                                                                                                          
    colbacktitle=green!25, coltitle=green!15!black,                                                                                                    
    boxsep=2pt, left=5pt, right=5pt, top=3pt, bottom=3pt,
    arc=2pt, outer arc=2pt                                                                                                                             
  ]               
  {\ttfamily\scriptsize                                                                                                                                
  \{\\            
  \hspace*{1em}"action": "USE\_SITE\_FIDELITY",\\
  \hspace*{1em}"rationale":\\
  \hspace*{2em}"Using site fidelity:\\
  \hspace*{2em}resource\_density=2.0 and robot is\\ 
  \hspace*{2em}set to use site fidelity; no active\\
  \hspace*{2em}pheromones to follow."\\
  \}                                                                                                                                                   
  }
  \end{tcolorbox}                                                                                                                                      
  \end{minipage}  
  \caption{LLM query at post-deposit decision in Fig.~\ref{fig:cluster-example}(c). The prompt carries the robot's local state, and the response specifies the chosen action together with a brief rationale.}                                                                   
  \label{fig:cluster-prompt}
  \end{figure}

We demonstrate the \textbf{qualitative} result with an example scenario, as in Fig.~\ref{fig:gazebo-cluster} with 4 robots in an arena sized at 6$\times$6m with 64 resources in clustered distribution, in which the LLM-Foraging makes decisions, as shown in Fig.~\ref{fig:cluster-example}. After the robot picks up a resource (Fig.~\ref{fig:cluster-example}a), it first returns to the central collection zone to deposit the resource (Fig.~\ref{fig:cluster-example}b). As described in Alg.~\ref{alg:llm_cpfa}, depositing resources will trigger a \emph{post-deposit} decision query through LLM (Fig.~\ref{fig:cluster-example}c). The query prompt and response from the LLM are shown in Fig.~\ref{fig:cluster-prompt}. No pheromone trails have been established yet, and the resource density is high at 2.0. Therefore, the LLM used site fidelity to perform an informed search of the site for additional resources. As shown in Fig.~\ref{fig:cluster-example}d, the robot goes back and picks more resources.

  \vspace{-0.4cm}

\subsection{Quantitative Results}
\begin{figure}
    \centering
    \includegraphics[width=1\linewidth]{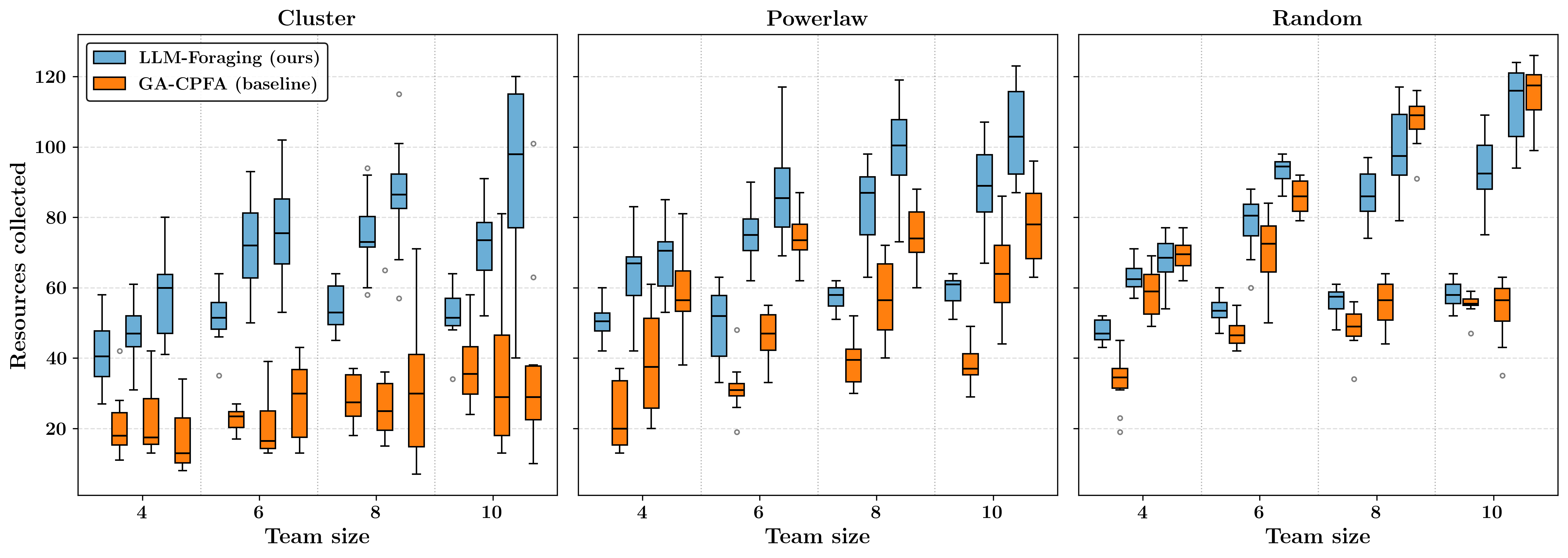}
    \vspace{-0.3cm}
    \caption{Resources deposited per trial across the 36 experimental combinations, with 10 trials each.
    Each panel shows one resource distribution.
    Within a panel, boxes are grouped by team size (4, 6, 8, 10), and the three sub-positions inside each group correspond to arenas of $6{\times}6$\,m (64 resources), $8{\times}8$\,m (128 resources), and $10{\times}10$\,m (256 resources).
    Blue boxes denote LLM-Foraging (ours), and orange boxes denote GA-CPFA (baseline).}
    \label{fig:boxplot}
\end{figure}

For the \textbf{quantitative} results, LLM-Foraging collects more resources than GA-CPFA in 33 of the 36 experimental combinations as described in (Fig.~\ref{fig:boxplot}).
The mean of per-combination relative improvements is +70\%, and the mean absolute gain is 22.9 resources per trial.
The three combinations in which GA-CPFA leads all fall under the $10{\times}10$\,m random condition and are discussed below.

The size of the LLM-Foraging advantage is related to the amount of spatial structure in the environment.
Averaged across team size and arena, the relative improvement is +142\% on clustered layouts, +49\% on powerlaw layouts, and +19\% on random layouts.
Clustered and powerlaw environments reward site fidelity and pheromone trails, which are the choices the LLM re-evaluates at each decision point.
Random layouts offer no exploitable structure, and both policies converge to comparable uninformed-search behavior.

LLM-Foraging deposits rise with team size across all nine arena-distribution combinations (Fig.~\ref{fig:boxplot}).
In the three combinations where GA-CPFA leads, the gap is at most 7\%.
Under random placement, the LLM's decision-point choices offer no improvement over GA-CPFA's fixed parameters.

Operational metrics of the LLM client match the per-decision cost envelope described in Section~\ref{sec-exp}.
Across the 360 trials, the controller issues 53{,}189 LLM calls with 3 fallbacks to the CPFA default.
Mean per-call latency is 6.6 seconds, with a maximum of 1024 tokens, using GPT-5-mini with low reasoning effort.
Because the three decision points fire only at event boundaries rather than at every control step, per-call latency is incurred sparsely over the 1200\,s trial horizon.

\section{Discussion and Conclusion}
\label{sec-conclusion}

We presented LLM-Foraging, a decentralized swarm controller that augments the CPFA state machine with an LLM tactical decision-maker at three structured decision points, namely \emph{post-deposit}, \emph{central-zone arrival}, and \emph{search starvation}. Each robot runs its own LLM client and queries it from locally observable state, while the existing CPFA motion and sensing stack executes the selected action. Because the LLM acts as a general decision policy rather than parameters fitted to one configuration, the controller is training-free at deployment and transfers across team size, arena size, and resource distribution without re-optimization.

From our experiments, GA-tuned CPFA does not scale to realistic deployment conditions, where every change in team size, arena, or resource layout would, in principle, require a fresh round of optimization. The baseline in our experiments was optimized on a single configuration and evaluated on 35 others, since per-condition returning would require roughly 21,600 hours at the cost given by Equation~\ref{eq:ga_eq}. The 70\% mean gap, therefore, reflects both algorithmic differences and the cost of re-optimization that GA-tuned controllers face under distribution shift. Under these conditions, LLM-Foraging is more consistent across configurations than GA-CPFA, and on understructured layouts, it correctly converges to behavior comparable with the baseline rather than over-committing to informed search when local cues are absent.

Limitations of the current evaluation include the latency introduced by each LLM call. In our case, the average latency is around 6.6\,s with GPT-5-mini. The LLM-Foraging framework itself is model-agnostic and accepts any LLM capable of producing the JSON response described in Section~\ref{sec-methods}. Because the three decision points fire only at event boundaries, this latency is incurred sparsely over the 1200\,s trial horizon, but it would be a binding constraint at finer decision granularity or on real hardware. A locally hosted small language model would reduce per-call latency and remove the dependency on external API availability.

\begin{credits}
\subsubsection{\ackname} The authors acknowledge the financial support provided by the NSF CISE Expand AI program (No. 2434916) and the NSF CISE MSI program (No. 2318682).

\end{credits}



%
%
%

\bibliography{references}
\bibliographystyle{splncs04}

\end{document}